\documentclass{sig-alternate}
\newfont{\mycrnotice}{ptmr8t at 7pt}
\newfont{\myconfname}{ptmri8t at 7pt}
\let\crnotice\mycrnotice%
\let\confname\myconfname%

\usepackage[german,american]{babel}
\usepackage[T1]{fontenc}
\usepackage[utf8]{inputenc}
\usepackage{times}
\usepackage{textcomp}
\usepackage{graphicx}
\usepackage{amssymb}
\usepackage{microtype} 

\usepackage{url}
\renewcommand{\tt}{\fontencoding{OT1}\fontfamily{cmtt}\selectfont}

\usepackage[usenames,dvipsnames]{color}
\usepackage{graphicx}
\usepackage{subcaption} 
\usepackage{booktabs}
\usepackage[numbers,sectionbib]{natbib}
\makeatletter
\def\@copyrightspace{\relax}
\makeatother

\begin{document}

\title{Clickbait Identification using Neural Networks}
\subtitle{The Whitebait Clickbait Detector at the Clickbait Challenge 2017}

\numberofauthors{1}
\author{
\alignauthor
Philippe Thomas\\
\affaddr{Deutsches Forschungszentrum f\"ur K\"unstliche Intelligenz, Germany}\\
\affaddr{philippe.thomas@dfki.de}\\
}

\maketitle

\begin{abstract}
This paper presents the results of our participation in the Clickbait Detection Challenge 2017.
The system relies on a fusion of  neural networks, incorporating different types of available informations.  
It does not require any linguistic preprocessing, and hence generalizes more easily to new domains and languages.
The final combined model achieves a mean squared error of 0.0428, an accuracy of 0.826, and a $\text{F}_1$ score of 0.564.
According to the official evaluation metric the system ranked 6th of the 13 participating teams.
\end{abstract}

\section{Introduction}
Clickbait refers to headlines of web content targeting the human ``curiosity gap''~\cite{Loewenstein94thepsychology}.
The reader is typically lured into clicking a target-link by raising interest into the advertised story mentioned in the teaser message, without providing enough details to satisfy the readers curiosity. 
Such clickbait-links often contain videos, picture galleries, or simple listings.
The content is mostly of little journalistic quality, but spreads well in social media by referring to soft topics.
Content describing such content (\emph{e.g.,} gossip, food news, or sensational stories) is often observed in tabloid newspapers.
The  conversion of a newspaper into tabloid format (also referred to as tabloidization) is often considered problematic~\cite{Rowe2011}.
However, there are also online magazines that provide clickbait titles on more serious topics.
According to an analysis all of the top 20 most prolific English news publishers on Twitter occasionally publish clickbait headlines~\cite{potthast:2016}.
Depending on the newspaper,  percentages of clickbait content ranges from 8\,\%  to an astonishing 51\,\%

In this publication we describe our approach in the Clickbait Detection Challenge 2017~\cite{potthast:2017a} to detect clickbait headlines using neural networks.

\section{Related Work}
Automated clickbait detection is a relatively recent task.
\citet{Chen:2015}  surveys potential methods and relevant concepts for the automatic detection of clickbait, including the existence of certain linguistic patterns to express clickbait headlines. 

\citet{BLOM201587} hypothesized that journalists use forward-referring headlines to increase click-rates. 
They analyzed 100,000 headlines from 10 different Danish news for forward-reference and observed that tabloidization seem to lead to a recurrent use of forward-reference.

\citet{Chakraborty2017} analyze the social sharing patterns of clickbait and non-clickbait tweets to determine the organic reach of such tweets.
To this end, the authors collected a number of twitter messages from newspaper accounts known to publish a high ratio of clickbait and non-clickbait content. 
The authors than examine differences between these two sets in terms of  consumer demographics,  follower graph structure, and type of text content.

\citet{potthast:2016} collected a clickbait corpus by sampling 150 tweets from each of the top 20 most prolific publishers on Twitter, totaling in 2992 tweets.
This contains several renowned newspapers, as well as publishers frequently associated with clickbaiting (such as BuzzFeed or Huffington Post).
All messages were rated being clickbait or not using the tweet text and the attached image. 
The authors also implement a clickbait detection model based on 215 features. 
This algorithm has been used as a baseline in the shared task. 
This dataset has been later extended by using crowdsourcing \cite{potthast:2017b}.

To the best of our knowledge previous works handled clickbait detection as binary classification task.
In contrast, the organizers of the Clickbait Detection Challenge 2017 proposed a regression problem, where the task is to judge the level of clickbaiting for a given tweet. 
Every tweet was annotated by five individual annotators into one of the four different classes: not click baiting (0.0), slightly click baiting (0.33), considerably click baiting (0.66), or heavily click baiting (1.0).
The annotations were provided as individual labels as well as different aggregation variables (\emph{i.e.,} mean, median, mode, and class).
The goal of the clickbait challenge was to accurately predict the mean value.

\section{Approach}
\begin{figure}[htbp]
  \centering
    \includegraphics[width=0.5\textwidth]{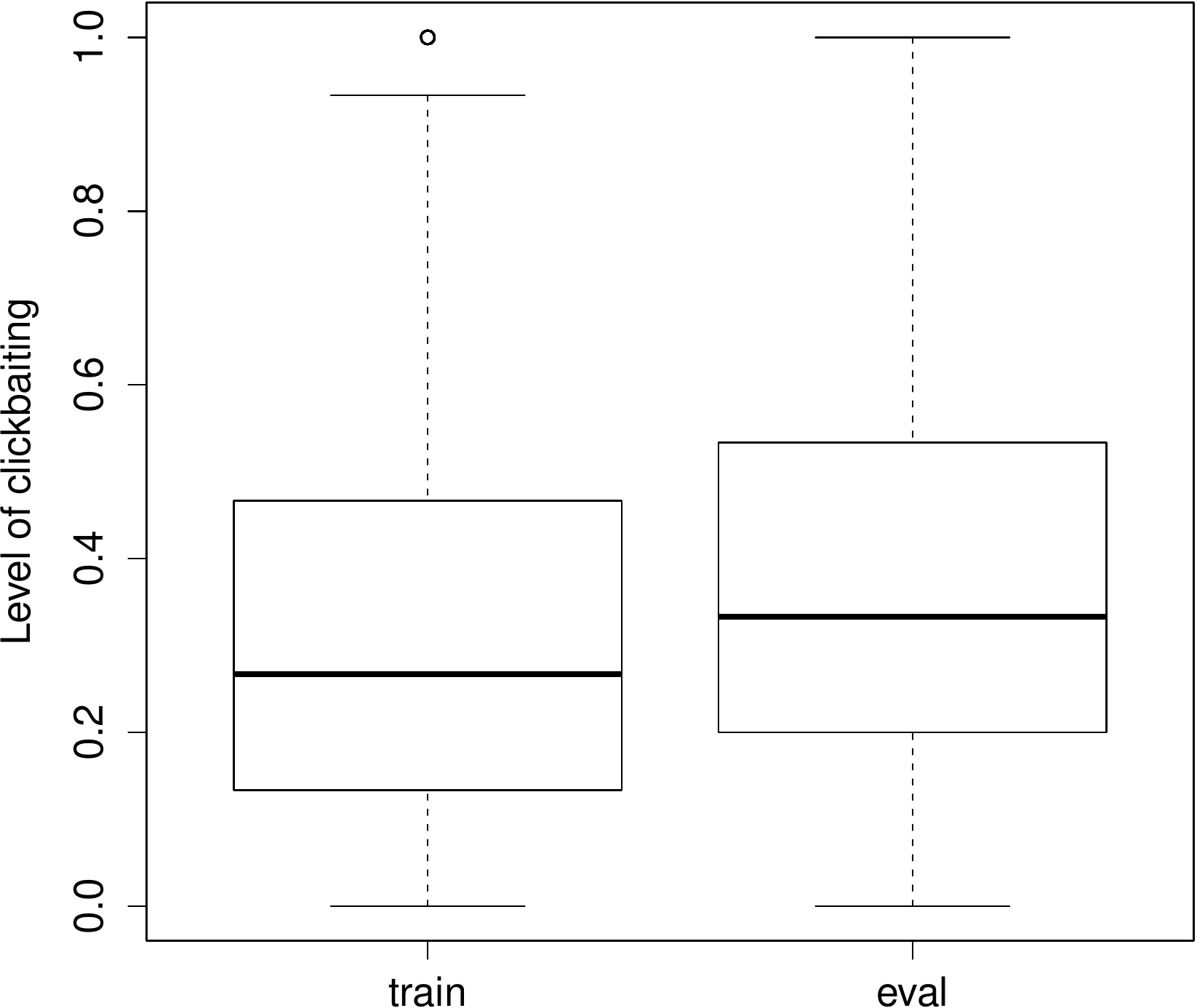}
    \caption{Distribution of clickbait scores on the two labeled datasets. }
 \label{fig:boxplot} 
\end{figure}

We used the Clickbait 2017 shared task data consisting of two labeled datasets, as described in Table~\ref{tab:statistics}.
For internal development purpose we used the larger dataset for training and the smaller dataset for evaluation.
The distribution of clickbait scores on these two datasets is shown in Figure~\ref{fig:boxplot} and indicates that there is a slightly higher proportion of clickbait articles in the eval-dataset. 
The difference between the two datasets is significant in terms of a Mann-Whitney-U test.
The official test dataset is hidden to the participants but the number of true labels was revealed after the competition.
Additionally, the organizers provided 80,012 unlabeled posts, which were not used in our approach. 

\begin{table}[htb]
\centering
\begin{tabular}{lrr}
\toprule
\multicolumn{1}{c}{\bf{Dataset}} & \multicolumn{1}{c}{\bf{clickbait}}  & \multicolumn{1}{c}{\bf{no-clickbait}} \\ 
 \cmidrule(lr){1-1} \cmidrule(lr){2-2} \cmidrule(lr){3-3} 
Train & 4,761 & 14,777 \\
Eval  & 762 & 1,697 \\
\midrule
Test   & 4,515 & 14,464 \\
\bottomrule
\end{tabular}
\caption{Statistics of the labeled datasets. Test data was not available to participants. }
\label{tab:statistics} 
\end{table}

Instances are provided as JSON objects and for each post we get a series of information.
For the \textit{post-tweet} this encompasses the text, potentially attached images, and the publication time.
For the \textit{target-article} this encompasses title, description, keywords, and paragraphs.
%

\begin{figure*}[htb]
    \centering
    \begin{subfigure}[b]{0.4\textwidth}
        \includegraphics[width=\textwidth]{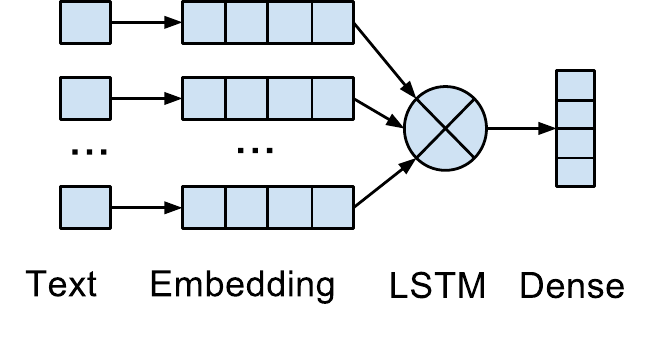}
        \caption{Example architecture used for textual data. Tokenized text is represented as word embeddings, which are then forwarded to a LSTM. Dropout and batch normalization is applied between individual layers.}
        \label{figure:net1}
    \end{subfigure}
    \hspace{5mm}
    \begin{subfigure}[b]{0.3\textwidth}
        \includegraphics[width=\textwidth]{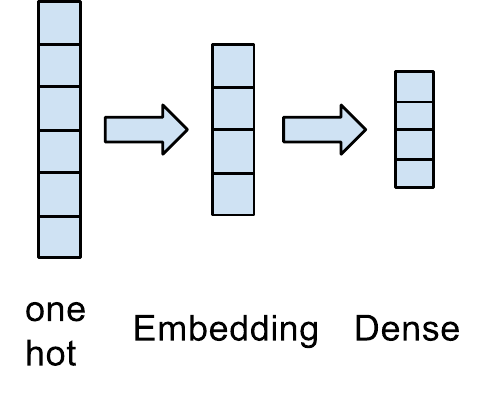}
        \caption{Example architecture used for categorical data. Categorical data is represented as one-hot encodings and internally converted to entity embeddings. }
        \label{figure:net2}
    \end{subfigure}
    \caption{Architectures for clickbait detection.}
\end{figure*}


\begin{figure*}
  \centering
    \includegraphics[width=1.0\textwidth]{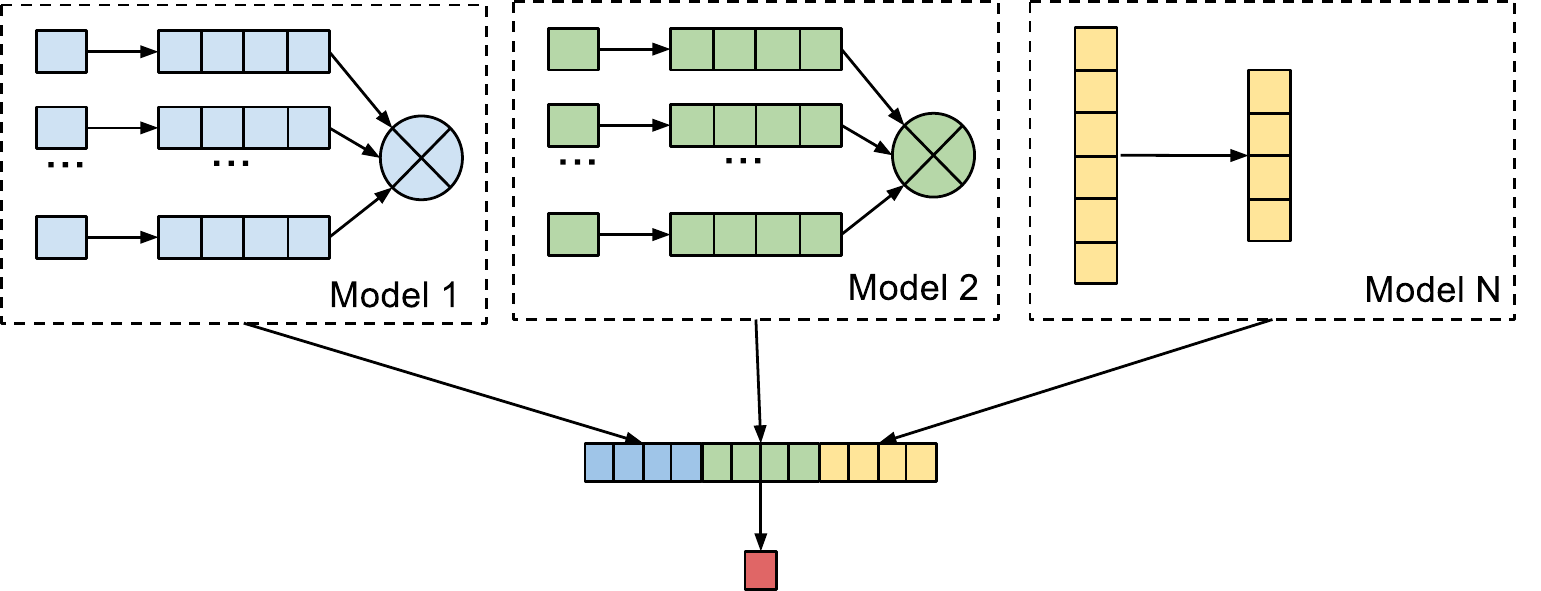}
    \caption{Neural network architecture using individual information sources. }
 \label{fig:finalNet} 
\end{figure*}

The clickbait detection approach follows closely the model used for geolocation prediction, described in \citet{Thomas2017}.
For text preprocessing, we use a simple whitespace tokenizer with lower casing, without any domain specific processing, such as unicode normalization~\cite{davis2001} or any lexical text normalization (see for instance \cite{Han:2011}).
The texts (post-text, target-title, target-paragraphs, target-description) are converted to word embeddings~\cite{MikolovSCCD13}, which are then forwarded to a Long Short-Term Memory (LSTM) unit~\cite{Hochreiter:1997}. 
In our experiments we randomly initialized embedding vectors.
We use batch normalization~\cite{Ioffe2015} for normalizing inputs in order to reduce internal covariate shift.
The risk of overfitting by co-adapting units is reduced by implementing dropout~\cite{Srivastava2014} between individual neural network layers.
An example architecture for textual data is shown in Figure~\ref{figure:net1}.
Post publication time (post-time) is  binned into hour ranges and than converted to one-hot encodings, which are forwarded to an internal embedding layer, as proposed by \citet{Guo16}. 
Again batch normalization and dropout is applied to avoid overfitting. The architecture is shown in Figure~\ref{figure:net2}.
We used RMSProp optimizer with early stopping and adaptive learning rate to train the neural networks.
For all parameters we did not perform any systematic optimization and used 100 embedding dimensions, batch size of 32, 100 training epochs, and a dropout rate of 0.3.

Finally, individual networks are fused by concatenating the dense output layers of the individual networks.
This concatenation is forwarded to a fully connected layer and used in our final model (see Figure~\ref{fig:finalNet} for the architecture).
Similar to \cite{Thomas2017}, we observed that the fusion of pre-trained models is beneficial in comparison to training a complete model from scratch.

\section{Evaluation Results}
Software has been uploaded to the TIRA experimentation platform~\cite{potthast:2014} for automatic evaluation of all teams participating in the shared task. 
The TIRA environment provides a uniform environment for participants to deploy and test submissions. 
Using TIRA, test data is not directly available to participants and therefore avoids the possibility of information leakage.

Results of different models on our evaluation corpus are shown in Table~\ref{tab:evaluation}.
According to our analysis, post-text is, in terms of mean squared error, the most productive information resource.
Using the fusion of individual neural networks we can successfully reduce the average error by 18\,\% (0.055 to 0.045) over the best individual model.

\begin{table}[htb]
\centering
\begin{tabular}{lrrrr}
\toprule
\multicolumn{1}{c}{\bf{Model}} & \multicolumn{1}{c}{\bf{MSE}}  &\multicolumn{1}{c}{\bf{MAS}} &\multicolumn{1}{c}{\bf{ACC}}& \multicolumn{1}{c}{\bf{$\textbf{F}_\textbf{1}$}} \\ 
\midrule
post-text 			& 0.055 & 0.153 & \bf{0.74} & \bf{0.50}\\
post-time 			& 0.057	& 0.19  & 0.69 & ---\\
target-paragraphs 	& 0.065 & 0.175 & 0.70 & 0.35\\
target-title  		& 0.066 & 0.168 & 0.70 & 0.41\\
target-description 	& 0.072 & 0.179 & 0.66 & 0.29\\
target-keywords 	& 0.073 & 0.194 & 0.68 & 0.20\\
\midrule
full model 			& \bf{0.045} & \bf{0.145} & \bf{0.74} &  0.39\\
\bottomrule
\end{tabular}
\caption{Performance ranked by mean squared error (MSE) on our evaluation corpus. Other metrics include mean absolute error (MAS), accuracy (ACC) and $F_1$-measure.}
\label{tab:evaluation} 
\end{table}

For the final submission we combined the two labeled datasets and retrained the neural networks using the same training regime.
The Whitebait clickbait detector achieved a mean squared error of 0.0428 and ranked 6th of the 13 participating teams. 
The  official results on the test corpus (0.0428) resembles closely the error rates observed on our internal evaluation corpus (0.045). 


\subsection{Incorporation of image information}
As the annotation process was supported by the image information, we assume that the teaser images might be helpful to predict the clickbait relevance of a given message. 
Also \citet{Ecker2014} state that images can be used to attract reader attention and are usually processed before the full article is read.
Therefore, we tried to incorporate the provided image information in the model.
In a first account we trained a small (3 layers) convolutional neural network from scratch.
Original sample size is increased using data augmentation, which modifies the original image using geometric and color augmentations~\cite{NIPS2012_4824}.
In our preliminary evaluations this model achieved random performance. 
In future experiments we would like to apply transfer learning from deep image classification models (\emph{e.g.,} VGG19~\cite{VGG19} or ResNet-50~\cite{ResNet}), trained on on the Image-Net dataset~\cite{imagenet}, to clickbait detection. 
These pre-initialized models should have already learned features, which might be relevant for our domain and should be less prone to overfitting to our data.

\section{Conclusion}
In this work we described our approach for the Clickbait Detection Challenge 2017.
We implemented a neural network, relying on different information resources.
The final combined model achieves a mean squared error of 0.0428, an accuracy of 0.826, and a $\text{F}_1$ score of 0.564.
In future work we would like to incorporate the image information into the final model.

\section*{Acknowledgments}
This  research  was  partially  supported  by  the  German  Federal  Ministry  of  Economics  and Energy   (BMWi)   through   the   projects  SD4M (01MD15007B) and SDW (01MD15010A) and by the German Federal Ministry of Education and Research (BMBF) through the projects ALL SIDES (01IW14002) and BBDC (01IS14013E).

\begin{raggedright}
\bibliography{clickbait17-notebook-lit}
\end{raggedright}
\end{document}